\documentclass[letterpaper, 10 pt, conference]{ieeeconf}

\IEEEoverridecommandlockouts                              
\usepackage{amsmath} % assumes amsmath package installed
\usepackage{amssymb}  % assumes amsmath package installed
\usepackage{graphicx}
\usepackage{hyperref}
\usepackage{siunitx}
\usepackage{subcaption}
\usepackage{booktabs}
\usepackage{xcolor}
\usepackage{textcomp}
\usepackage{gensymb} %for degree symbol
\usepackage{multirow} %for tables
\usepackage{comment}
\usepackage{verbatim}
\usepackage[skip=0pt, font=footnotesize]{caption}
\usepackage{mathtools}
\usepackage{bm}
\usepackage[noadjust]{cite}
\usepackage{makecell}
\usepackage{tabularx}
\usepackage{pifont}
\usepackage{layouts}
\newcommand{\cmark}{\ding{51}}%
\newcommand{\xmark}{\ding{55}}%
\newcolumntype{M}[1]{>{\centering\arraybackslash}m{#1}}
\makeatletter
\def\endthebibliography{%
	\def\@noitemerr{\@latex@warning{Empty `thebibliography' environment}}%
	\endlist
}
\makeatother
\makeatletter 
\newcommand\Label[1]{&\refstepcounter{equation}(\theequation)\ltx@label{#1}&}
\makeatother
\makeatletter
\let\@oldmaketitle\@maketitle% Store \@maketitle
\renewcommand{\@maketitle}{\@oldmaketitle% Update \@maketitle to insert...
	\centering
	\includegraphics[width=\textwidth]{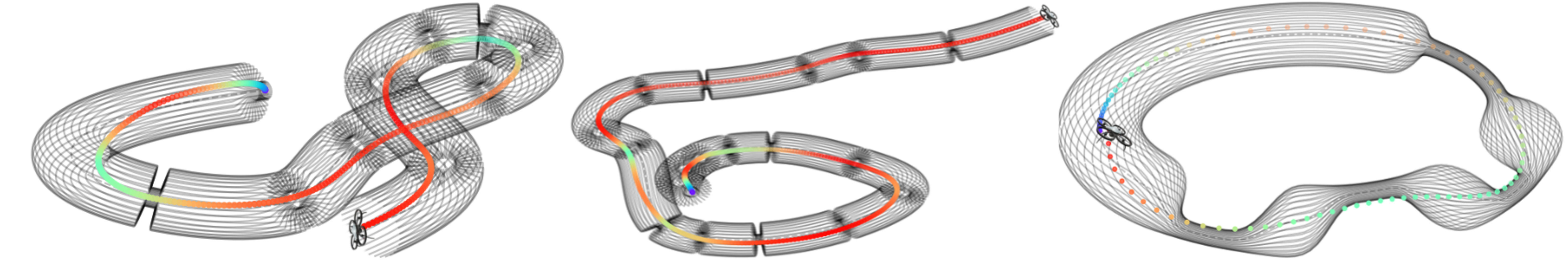}
	\vspace{0.2mm}
	\captionof{figure}{A quadrotor navigates at near-time-optimal performance through three different dynamic tunnels: \emph{AlphaPilot} (left), \emph{AirSim} (center) and \emph{TubeTwist} (right). The respective motions have been computed by our NMPC based real-time control method capable of handling time-variant spatial bounds.}
	\label{fig:trajectory_fancy}
	\vspace{-4.5mm}
	}% ... an image
\makeatother
\title{\LARGE \bf
Towards Time-Optimal Tunnel-Following for Quadrotors}

\author{Jon Arrizabalaga$^{1}$ and Markus Ryll$^{1}$% <-this % stops a space
	\thanks{$^{1}$Autonomous Aerial Systems Lab, Department of Aerospace and Geodesy,  Technical University of Munich, Germany. E-mail: {\tt\small jon.arrizabalaga@tum.de} and {\tt\small markus.ryll@tum.de}
	}%
}
\IEEEoverridecommandlockouts

\usepackage{tikz}
\usepackage{textcomp}
\usepackage{hyperref}
\usepackage{lipsum}

\newcommand\copyrighttext{%
	\footnotesize Published in IEEE International Conference on Robotics and Automation (ICRA), Philadelphia, USA, May 2022.\newline
	 \textcopyright 2022 IEEE. Personal use of this material is permitted.
	Permission from IEEE must be obtained for all other uses, in any current or future media, including reprinting/republishing this material for advertising or promotional purposes, creating new collective works, for resale or redistribution to servers or lists, or reuse of any copyrighted component of this work in other works.}
\newcommand\copyrightnotice{%
	\begin{tikzpicture}[remember picture,overlay]
		\node[anchor=south,yshift=10pt] at (current page.south) {\fbox{\parbox{\dimexpr\textwidth-\fboxsep-\fboxrule\relax}{\copyrighttext}}};
	\end{tikzpicture}%
}
\begin{document}
\begin{comment}
\bstctlcite{IEEEexample:BSTcontrol} %avoid dashing out same authors. See IEEEtrans_bst_HOWTO.pdf
\end{comment}
\maketitle
\copyrightnotice
%\thispagestyle{empty}
%\pagestyle{empty}
%textwidth in cm: \printinunitsof{cm}\prntlen{\textwidth}
\begin{abstract}
Minimum-time navigation within constrained and dynamic environments is of special relevance in robotics. Seeking time-optimality, while guaranteeing the integrity of time-varying spatial bounds, is an appealing trade-off for agile vehicles, such as quadrotors. State-of-the-art approaches, either assume bounds to be static and generate time-optimal trajectories offline, or compromise time-optimality for constraint satisfaction. Leveraging nonlinear model predictive control and a path parametric reformulation of the quadrotor model, we present a real-time control that approximates time-optimal behavior and remains within dynamic corridors. The efficacy of the approach is evaluated by simulated results, showing itself capable of performing extremely aggressive maneuvers as well as stop-and-go and backward motions.
\end{abstract}
\textbf{Video}: \url{https://youtu.be/Apc8MCu7Yvo}
% Keywords: mobile robotics, motion planning, model-predictive-control, caster-wheel

%\the\textwidth
%\vspace{-3mm}
\section{INTRODUCTION}
\noindent Autonomous robotic systems are deployed in dynamic environments, where conditions and obstacles are time-variant. Successfully navigating under these circumstances, implies remaining within a safe and time-variant corridor \cite{siegwart2011introduction}. Time-optimality within dynamic environments poses a particular challenge, since the minimum-time trajectory changes along with the environment.  
 
Because of their inherit agility \cite{kaufmann2020deep}, quadrotors are frequently used in space-constrained and time-critical operations, such as delivery, surveillance, inspection, and search-and-rescue \cite{shakhatreh2019unmanned, kumar2012opportunities}. These are distinguished by the necessity of being time-optimal and subjected to strong environmental changes. Pushing the physical limits of quadrotors in unknown variant environments brings up a twofold challenging trade-off: 1) quadrotors are underactuated systems, i.e., longitudinal and angular dynamics are coupled, 2) the motion aggressiveness is bounded by the dynamicity of the environment. In other words, time-optimal performance in dynamic environments consists on \emph{finding the optimal trade-off between longitudinal and angular accelerations (1) in accordance with variant spatial bounds (2)}. 

Previous works have decoupled these two features, either by assuming the environment to be invariant and tracking an offline-generated minimum-time trajectory \cite{foehn2021time, spedicato2017minimum}, limiting time-optimality to waypoint following and neglecting spatial bounds \cite{romero2021model}, or compromising time-optimality due to actuation limit relaxations \cite{mellinger2011minimum, richter2016polynomial, van2013time, tordesillas2019faster} and collision-free constraints \cite{chen2016online}.

Given its recent applications on fast dynamical systems and its support for nonlinear system models \cite{diehl2002real, neunert2016fast}, Nonlinear Model Predictive Control (NMPC) is a well-suited framework for motion planning of quadrotors. NMPC can find the (locally) optimal control action considering a nonlinear cost function while incorporating constraints on nonlinear functions of states and inputs \cite{camacho2013model}. Therefore, NMPC can account for time-varying spatial bounds, while enforcing constraints in the true physical limits, i.e. thrust commands, and thus, ensuring near-time-optimality of the computed solution. 

Multiple research challenges arise, including how to construct a lightweight NMPC capable of running in real-time and how its performance compares to the previously mentioned offline approaches.

 \subsection*{Contributions}
\noindent We present an NMPC-based real-time control that computes near-time-optimal quadrotor trajectories within the free space associated with a dynamic environment -- denoted as \emph{tunnel}. To the best of the authors' knowledge, this is the first real-time approach tackling time-optimality within constrained and time-variant spaces in quadrotors. In contrast to state-of-the-art methods, our solution is immediately deployable, without requiring prior environmental knowledge or offline computation, and is capable of ranging from extremely aggressive maneuvers to stopping and even reversing.

Our method is comprised of two main ingredients: 1) we perform a path-parametric reformulation of the quadrotor system dynamics, allowing to embed path properties and variant spatial constraints effectively in the optimization and 2) by solely bounding thrust commands, the NMPC is empowered to reach the true physical limits of the system.
\begin{comment}
and 3) the differential flatness property is exploited to relax the minimum sampling time, by mapping computed thrust commands to angular rates, without loss of time-optimality.
\end{comment}
 
According to empirical results, the suggested approach 1) approximates time-optimal behavior, while adapting its speed depending on the shape and size of the tunnel 2) is agnostic to tunnel changes beyond the prediction horizon, 3) can stop or navigate backwards when abrupt tunnel modifications occur within the horizon and 4) shows convergence to the tunnel if a given modification is excessively aggressive.

The remainder of this paper is structured as follows: Section~\ref{sec:related_work} provides more details on quadrotor motion planning and spatial reformulation of dynamical systems. Section~\ref{sec:plant_model} presents the path-parametric reformulation of the quadrotor dynamics. Section~\ref{sec:nmpc_formulation} exemplifies the complete NMPC formulation, including the tunnel's parametrization. Experimental setup and results are shown in Section~\ref{sec:experiments} before Section~\ref{sec:conclusion} presents the conclusions.

\section{RELATED WORK}\label{sec:related_work}

\subsection{Time-optimal and collision-free planning}
\noindent 
Path planning algorithms for quadrotors can be classified into two variants based on the state-space of the planning. Exploiting the property of differential flatness, all quadrotor states can be mapped into four differentially flat outputs. By approximating them to piecewise constant polynomials and imposing continuity constraints in the respective derivatives, smooth trajectories can be generated. Since this concept was presented in \cite{mellinger2011minimum}, further improvements to solve the underlying quadratic program have allowed for extremely fast computations \cite{richter2016polynomial,burke2020generating}. This has played a key role in standard decoupled approaches like \cite{chen2016online}, where a collision-free path provided by a global planner is further refined by a local planner running at a higher sampling rate.

The computational advantages brought by the flatness property and its respective polynomials come at the expense of 1) highly non-linear mappings from the flat outputs to thrust commands and 2) non-time-optimality inherited from the minimization of polynomials. The first suggests that applying differential flatness prevents setting the thrust constrains required for reaching physical limits. The second implies that minimizing polynomials is in direct opposition to maximizing acceleration. 

Therefore, methods that account for the entire state-space are better suited to converge to minimum-time trajectories. Focusing on time-optimality when navigating through a set of locations, \cite{foehn2021time} formulated an optimization problem that parametrizes progress with dynamic waypoints through Complementary Progress Constraints (CPC). Computed paths prove to be theoretically optimal according to the physical limits of the quadrotor and capable of outperforming expert pilots. Tackling the same problem, \cite{song2021autonomous} presents a Deep Reinforcement Learning based algorithm that calculates near-time-optimal solutions while dealing with uncertainty in waypoint poses. Despite these methods allow for a close approximation of the true time-optimal path, they are limited to offline computations, and thus, are intractable for real-time applications. This burden is alleviated in \cite{romero2021model}, where a Contouring Control MPC approach inspired by \cite{10.1007/978-3-030-71151-1_4} and \cite{brito2019model} allows for near-time-optimal performance in real-time. However, the proposed approach is limited to waypoint following and does not account for spatial bounds. Moreover, the underlying gradient in the Contouring Control necessitates additional constraints, compromising the solution’s versatility and applicability for dynamic scenarios, such as when reversal motions are required. 

In contrast, our approach runs in real-time, requires no prior knowledge of the environment, and is capable of performing aggressive maneuvers as well as stopping and navigating backwards. These features are attributed to a coordinate transformation that enable a more precise embedding of the environment's geometric properties into the NMPC's underlying optimization problem. 

\subsection{Path-parametric reformulation}
\noindent Benefits of transforming time-dependent dynamics into spatial-dependent were first presented for robot systems in \cite{pfeiffer1987concept, verscheure2009time}. Extensions of this reformulation to planar vehicles proved its ability to trade-off between reference tracking and obstacle avoidance \cite{gao2012spatial, frasch2013auto}. Leveraging this reparameterization  and exploiting advances in embedded optimization solvers, \cite{verschueren2014towards} developed a real-time NMPC for miniature racing cars that could approximate time-optimal performance. In \cite{kloeser2020nmpc}, further modifications in the spatial representation of the plant model allowed including online obstacle avoidance, while still being near-time-optimal. Focusing on robot manipulators, \cite{van2016path} extended the path-parametric reformulation to all three dimensions and presented an NMPC capable of trading-off between tracking accuracy and speed.

In the context of quadrotors, by combining path-parametrization with differential flatness, \cite{van2013time} presented an efficient optimization problem for time-optimal trajectory planning of two-dimensional models in cluttered environments. In \cite{kumar2017path} the spatial reformulation was implemented to all three dimensions, allowing to decouple the tangential and transverse dynamics of the quadrotor. Similarly, \cite{spedicato2017minimum} benefited from a three-dimensional reparameterization for offline computation of minimum-time and collision-free trajectories.

Previous work suggests that, in contrast to planar vehicles, the realm of quadrotors has not fully leveraged the capabilities inherited in path-parametrization. We deem to close this gap by combining time-optimality techniques from the aforementioned two-dimensional solutions with a complete spatial conversion of the quadrotor dynamics. 

\section{PATH-PARAMETRIC MODEL}\label{sec:plant_model}

\subsection{Quadrotor Model}
\noindent The quadrotor is modeled as a six degree-of-freedom rigid body. Its states are described in the world-frame with $\bm{x} = \left[\bm{p}_{\text{W}},\bm{v}_{\text{W}},\bm{q}_{\text{W}},\bm{\omega}_{\text{W}}\right]$, where $\{\bm{p}_{\text{W}},\bm{v}_{\text{W}},\bm{\omega}_{\text{W}}\} \in \mathbb{R}^3$ refer to the position, velocity and body rates, while the attitude is represented by a unit quaternion $\bm{q}_\text{W} \in \mathbb{R}^4$. The collective thrust and torques in  body-frame are the inputs with $\{\bm{f}_\text{B},\bm{\tau}_{\text{B}}\} \in \mathbb{R}^3$. Since frame subscripts remain constant, they will be dropped. The dynamics of the system are as common

\pagebreak
\begin{subequations} %\label{eq:nominal_model}
\begin{align*}
	 \bm{\dot{p}}&=\bm{v} &\Label{eq:pdot}\quad
	 \bm{\dot{v}}&=\bm{g} + \frac{1}{m}\text{R}(\bm{q})\bm{f}&\Label{eq:vdot} \\
	 \bm{\dot{q}}&= \frac{1}{2}\Lambda(\bm{\omega})\bm{q}&\Label{eq:qdot}\quad  \bm{\dot{\omega}} &= \text{J}^{-1} \left(\bm{\tau} - \bm{\omega} \times \text{J}\,\bm{\omega} \right)&\Label{eq:rdot}
\end{align*}
\noindent with $m$, $\text{J}$ and $\bm{g}$ being the quadrotor's mass, moment of inertia matrix and gravity, while $\Lambda$ and $\text{R}$ refer to the skew-symmetric matrix of the body rates and rotation matrix of the quaternion. Body forces $\{\bm{f},\bm{\tau}\}$ can be further mapped to single rotor thrust commands $\bm{\eta} = \left[\eta_1, \eta_2, \eta_3, \eta_4\right]$ according to
\begin{equation*}
\begin{aligned}
	&\bm{f} = \begin{bmatrix}
	0\\0\\\sum_{i=1}^{4}\eta_i
	\end{bmatrix}
\end{aligned}
\text{and}\quad
\begin{aligned}
&\bm{\tau} = \begin{bmatrix}
l/\sqrt{2} (\eta_1+\eta_2-\eta_3-\eta_4)\\
l/\sqrt{2} (-\eta_1+\eta_2+\eta_3-\eta_4)\\
c_{\tau} (\eta_1-\eta_2+\eta_3-\eta_4)\\
\end{bmatrix}
\end{aligned}	
\end{equation*}
\end{subequations}
where $l$ is the arm-length and $c_\tau$ the force-torque mapping constant of the quadrotor.

\subsection{Path-Parametric Reformulation}

\noindent To embed the path characteristics into the system dynamics, we reformulate the equations of motions for position \eqref{eq:pdot} and velocity \eqref{eq:vdot}. This is done in two steps: first we describe the path using Frenet-Serret's theorem, and then perform a spatial reformulation. When doing so, we will combine ideas from \cite{spedicato2017minimum} and \cite{van2016path}.

As a starting point, we introduce a \emph{progress variable} $s$ exemplifying the arc-length of a curve and a $C^\infty$ continuous \emph{progress-function} $\bm{\gamma}(s)$ that maps $s$ to its respective three-dimensional position in the Euclidean space. Thus, the path can be defined as $\Gamma = \{\bm{\gamma}(s)\in\mathbb{R}^3\ |\ s \in[0,L]\}$.

Discretizing $\Gamma$ into $M$ points and assigning each one an orthonormal basis in the Frenet-Serret coordinate system allows for describing the pose as a function of the path's curvature $\kappa$, torsion $\sigma$ and progress-function $\bm{\gamma}$ (see Fig.~\ref{fig:spatial_coordinates}). These are related by Frenet-Serret's theorem \cite{kuhnel2015differential}:
\begin{subequations}\label{eq:frenet-serret}
\begin{gather}
	\bm{\mathfrak{t}} = \bm{\gamma}^\prime,\quad
	\bm{\mathfrak{n}} = \bm{\gamma}^{\prime\prime}/||\kappa||_2\,,\quad
	\bm{\mathfrak{b}} = \bm{\mathfrak{t}} \times \bm{\mathfrak{n}},\\
	\bm{\mathfrak{t}}^{\prime} = \kappa\,\bm{\mathfrak{n}}, \quad
	\bm{\mathfrak{n}}^{\prime} = -\kappa\,\bm{\mathfrak{t}}+\sigma\,\bm{b}\,,\quad
	\bm{\mathfrak{b}}^{\prime} = -\sigma\,\bm{\mathfrak{n}}
\end{gather}
\end{subequations}
where $\bm{\mathfrak{t}}$, $\bm{\mathfrak{n}}$ and $\bm{\mathfrak{b}}$ are the tangent, normal and binormal components of the Frenet-Serret frame $\{\text{FS}\}$ and $(\cdot)^{\prime}$ represents the derivative with respect to the progress variable $s$. Notice that all terms in \eqref{eq:frenet-serret} are continuous and uniquely dependent on $s$, and thus, we can approximate them by third order B-splines. 
\begin{figure}
	\centering
	\captionsetup{type=figure}\addtocounter{figure}{-1}%PROVISIONAL
	\includegraphics[width=\linewidth]{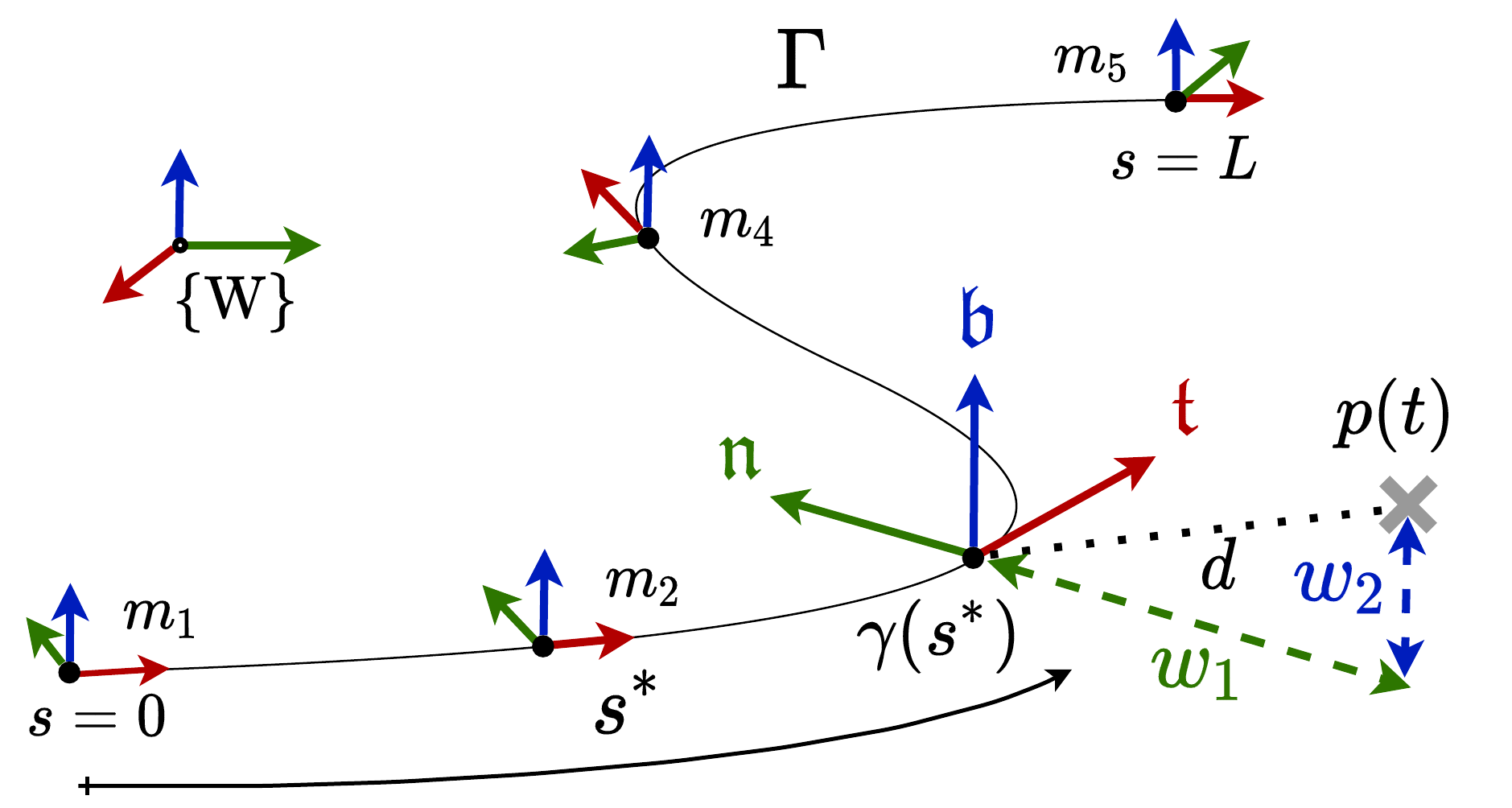}
	\caption{Path representation according to progress variable $s$ and Frenet-Serret frames. The distance $d$ between the quadrotor $p(t)$, depicted as a gray cross, and the closest point on the path $\gamma(s^*)$ is projected onto the transverse coordinates $w_1$ and $w_2$. For readability, the illustration assumes the torsion along the entire path to be zero.}\label{fig:spatial_coordinates}
\end{figure}

Putting together all three components, the orientation along the path is defined by the Frenet-Serret rotation matrix $\text{R}_{\text{FS}}(s) = \left[\bm{\mathfrak{t}}(s),\bm{\mathfrak{n}}(s),\bm{\mathfrak{b}}(s)\right]$. Assuming that the quadrotor is located in position $\bm{p}(t)$, the progress of the closest point on the path $\Gamma$ is defined as 
\begin{equation}\label{eq:s_closest_ocp}
s^*(t) = \text{arg}\min_{s}\frac{1}{2}||\bm{d}_\text{W}(t)||,\quad 
\bm{d}_\text{W}(t) = \bm{p}(t)-\bm{\gamma}(s)\;.
\end{equation}

From now onward the dependency of $s^*(t)$ on time will be dropped. Translating the distance vector to the Frenet-Serret frame associated to the closest point defines $\bm{d}_\text{FS}(t) = \text{R}_{\text{FS}}(s^*)^\intercal \bm{d}_\text{W}(t)\,,$ whose first row, i.e. the tangent component, is zero. Since the remaining two elements are the perpendicular projections of the distance from path $\Gamma$, they will be named \emph{transverse coordinates} $\bm{w}=\left[w_1,w_2\right]$. Consequently, the distance is equivalent to $\bm{d}_\text{FS}(t) = \left[0,w_1(t),w_2(t),\right]$ with
\begin{equation}\label{eq:transverse_coordinates}
w_1(t) = \bm{\mathfrak{n}}(s^*)\bm{d}_{\text{W}}(t), \; 
w_2(t) = \bm{\mathfrak{b}}(s^*)\bm{d}_{\text{W}}(t)\;.
\end{equation}
Either from applying the first optimality condition to \eqref{eq:s_closest_ocp} as in \cite{van2016path} or from rewriting equation \eqref{eq:pdot} with the transverse coordinates and derivating with respect to time as in \cite{spedicato2017minimum}, the velocity of the progress respective to the closest point on path $\Gamma$ can be expressed as
\begin{subequations}\label{eq:spatial_dot}
\begin{equation}\label{eq:sdot}
\dot{s}^*(t) = \frac{\bm{\mathfrak{t}}(s^*)^\intercal\bm{v}(t)}{1-\kappa(s^*) w_1(t)}
\end{equation}
and combining it with \eqref{eq:transverse_coordinates} leads to
\begin{gather}
\dot{w}_1(t) = \bm{\mathfrak{n}}(s^*)\bm{v}(t)+\sigma(s^*)\dot{s}^*(t)w_2(t)\,,\label{eq:trasnverse_coordinate_dot1}\\
\dot{w}_2(t) = \bm{\mathfrak{b}}(s^*)\bm{v}(t)-\sigma(s^*)\dot{s}^*(t)w_1(t)\,.\label{eq:trasnverse_coordinate_dot2}
\end{gather}
\end{subequations}
 For a detailed derivation of \eqref{eq:trasnverse_coordinate_dot1} and \eqref{eq:trasnverse_coordinate_dot2}, please refer to \cite{spedicato2017minimum}. The singularity in \eqref{eq:sdot} can be circumvented by requiring $1-\kappa(s^*) w1(t) > 0$, or equivalently $r(s^*) > w1(t)$, i.e. the normal component of the distance between the quadrotor and the path needs to be smaller than the radius of the curve.
 
 The equations of motion for the tangential and transverse coordinates in \eqref{eq:spatial_dot} can already replace the position dynamics in \eqref{eq:pdot}. However, to fully embed the path properties within the system dynamics, the integration chain of the path coordinates should be extended up to the degree where the input forces appear. In the case of a quadrotor, the collective thrust arises in the longitudinal acceleration \eqref{eq:vdot}. Therefore, we are interested in differentiating the path coordinates up to the second order. This is done by taking the time derivatives of the equations in \eqref{eq:spatial_dot}:
 \begin{subequations}\label{eq:spatial_dotdot}
 	\begin{gather}
 	\ddot{s}^*(t) = \frac{\bm{\mathfrak{t}}\dot{\bm{v}}+\dot{\bm{\mathfrak{t}}}\bm{v}}{1-\kappa w_1} + \bm{\mathfrak{t}}\bm{v}\left[\frac{w_1\dot{\kappa}+\dot{w_1}\kappa}{(1-\kappa w_1)^2}\right]\\
 	\ddot{w}_1(t) = \bm{\mathfrak{n}}\dot{\bm{v}}+\dot{\bm{\mathfrak{n}}}\bm{v}+\dot{\sigma}\dot{s}^*w_2+\sigma\ddot{s}^*w_2+\sigma\dot{s}^* \dot{w}_2\\
 	\ddot{w}_2(t) = \bm{\mathfrak{b}}\dot{\bm{v}}+\dot{\bm{\mathfrak{b}}}\bm{v}-\dot{\sigma}\dot{s}^*w_1-\sigma\ddot{s}^*w_1-\sigma\dot{s}^* \dot{w}_1
 	\end{gather}
 \end{subequations} 
 where dependencies have been omitted for clarity. The time derivatives of curvature, torsion and Frenet-Serret coordinates can be calculated according to the chain rule  $\dot{(\cdot)}= \frac{\partial (\cdot)}{\partial s}\frac{\partial s}{\partial t}=(\cdot)^\prime\dot{s}^*$, while the longitunal acceleration $\bm{\dot{v}}$ is known from \eqref{eq:vdot}. The velocity is the only variable that remains for being translated to spatial coordinates. To this end we reformulate the position as $\bm{p}(t) = \bm{\gamma}(s^*)+\text{R}_\text{FS}(s^*)\bm{d}_{\text{FS}}(t)$ and differentiate with respect to time:
 \begin{equation}\label{eq:spatial_v}
 \bm{v}(t) = \dot{\bm{\gamma}}+\dot{\text{R}}_\text{FS}\,\bm{d}_{\text{FS}}+\text{R}_\text{FS}\,\dot{\bm{d}}_{\text{FS}}
 \end{equation}
 whose terms can be obtained by applying the aforementioned chain rule. Finally, by combining \eqref{eq:vdot} with \eqref{eq:spatial_dotdot} and \eqref{eq:spatial_v}, the euclidean position and velocity states can be replaced by the progress variable, transverse coordinates and its respective derivatives, i.e. $\bm{x} = \left[s^*,\bm{w},\dot{s}^*,\dot{\bm{w}},\bm{q},\bm{\omega}\right]$. From here on to simplify the notation $s^*$ will be referred to as $s$.
 
\section{PROGRESS-MAXIMIZATION NMPC} \label{sec:nmpc_formulation}
\noindent The optimal control problem (OCP) addressing minimum-time navigation within dynamic corridors is formulated as:
\begin{subequations} \label{eq:OCP}
\begin{alignat}{3}
&\min_{\bm{x}(.), \bm{u}(.),T} T\\ 
&\quad\text{s.t.}&&\bm{x}_{0} = \bm{x}_{T} \\
&&&T\geq 0\\
&&&\dot{\bm{x}}(t) = f(\bm{x}(t),\bm{u}(t)), &\quad& t \in \left[0,\,T\right]\\
&&&\underline{\bm{u}} \leq \bm{u}(t) \leq \overline{\bm{u}}, &\quad& t \in \left[0,\,T\right]\\
&&& \phi\left(\bm{w}(t),\xi(t)\right) \leq 0 &\quad& t \in \left[0,\,T\right] \label{eq:tunnel_constraint_time}
\end{alignat}
\end{subequations}
where $\bm{x} = \left[s,\bm{w},\dot{s},\dot{\bm{w}},\bm{q},\bm{\omega}\right]$ and $\bm{u} =\left[\bm{\eta}\right]$ refer to the quadrotor's states and controls, while $f$ is the first-principles-based model derived in Section \ref{sec:plant_model}. Other than bounding thrust commands by $\underline{\bm{u}}$ and $\overline{\bm{u}}$, we limit the navigation space to a time-varying set $\xi(t)$ according to function $\phi$.

To approach online time-optimal navigation, we approximate the OCP \eqref{eq:OCP} by a Nonlinear Program (NLP) according to the multiple-shooting approach \cite{bock1984multiple} in which the optimization horizon $T$ is split into $N$ sections with constant decision variables. Similarly to \cite{verschueren2014towards} and \cite{kloeser2020nmpc}, by tracking a slightly infeasible reference $s_\text{ref}$ and its respective velocity $\dot{s}_\text{ref}$, we incite time-optimal behavior. For a given progress velocity, the respective arc-length can be defined as $s_{\text{ref},k} = s_0 + \frac{\dot{s}_\text{ref,k}}{T} \frac{k}{N}$ for $k = 0,\cdots,N$, and thus $\bm{x}_{\textbf{ref}} = \left[s_\text{ref},\bm{c_w}(s_\text{ref}),\dot{s}_\text{ref},\bm{0}\right]$, where $\bm{c_w}$ refers to the transverse coordinates of the tunnel's center line.

To leverage the computational advantages of the Gauss-Newton Hessian approximation, the cost function is implemented in a quadratic least-squares fashion. When doing so, the tractable NLP that is solved at every NMPC iteration and approximates OCP \eqref{eq:OCP} is given by:
\begin{subequations}\label{eq:NLP}
	\begin{equation}
	\min_{\substack{\bm{x}_{\,0},\cdots,\,\bm{x}_{\,N},\\\bm{u}_{\,0},\cdots,\,\bm{u}_{\,N}}}
	\sum_{k=0}^{N-1}
	\left|\left|\bm{x}_{k}-\bm{x}_{\textbf{ref},\,k}\right|\right|_{Q}^2+
	\left|\left|\bm{u}_{\,k}\right|\right|_{R}^2 +
	\left|\left|\bm{x}_N-\bm{x}_{\textbf{ref},\,N}\right|\right|_{Q_N}^2 
	\end{equation}
	\begin{alignat}{3} 
	\text{s.t.}\quad &\bm{x}_{\,0} = \bm{x}_{\,c}\\
	&\bm{x}_{\,k+1} = F(\bm{x}_{\,k},\bm{u}_{\,k}, \Delta\,t), &\quad&k = 0,\cdots,\,N-1\\
	&\underline{\eta} \leq \eta_{i,\,k} \leq \overline{\eta}, &\quad&\substack{k = 0,\cdots,\,N-1,\\i = 1,\cdots,4}\\
	&\phi\left(\bm{w}_{\,k},\xi(s_k)\right) \leq 0,    &\quad&k = 0,\cdots,\,N-1 \label{eq:tunnel_constraint}
	\end{alignat}
\end{subequations}
where $\bm{x}_{\,c}$ is the quadrotor's current state, $F$ is a numerical integrator for the spatial quadrotor dynamics and $Q,R,Q_N$ are weighting matrices. Control commands are softened by appending rotor-thrusts to the state vector $\bm{x} = \left[s,\bm{w},\dot{s},\dot{\bm{w}},\bm{q},\bm{\omega},\bm{\eta}\right]$ and assigning their respective derivatives to the system's inputs $\bm{u} =\left[\dot{\bm{\eta}}\right]$. To guarantee feasibility, constraint \eqref{eq:tunnel_constraint} is relaxed according to slack variables that are linearly and quadratically penalized in the cost function. 

As stated in \eqref{eq:tunnel_constraint_time} and \eqref{eq:tunnel_constraint}, to ensure that the quadrotor stays within an arbitrary geometry, we define a function $\phi$. The loss of curvature inherited from the linearization of non-linear constraints might compromise convergence in the optimization problem \cite{verschueren2016exploiting}. Therefore, when formulating spatial bounds in $\phi$, linear representations are preferred over non-linear ones. Without loss of generality, we approximate the tunnel to a polyhedron of $n_p$ sides, whose inner space can be represented as the following convex set:
\begin{subequations}
\begin{equation*}
\Omega = \{s,w_1,w_2\;|\;s \in [0,L] \land \;\bm{a}(s) w_1 + \bm{b}(s) w_2 < \bm{c}(s)\} \;, 
\end{equation*}
where $\{\bm{a}(s),\bm{b}(s),\bm{c}(s)\}\in \mathbb{R}^{n_p}$ describe the polyhedron at a given progress $s\in[0,L]$. Putting the inequality in the form of \eqref{eq:tunnel_constraint}, results in
\begin{equation*} \label{eq:tunnel_constraint_1}
\phi :=\xi(s)\begin{bmatrix}\bm{w}\\-1\end{bmatrix} \leq \bm{0} \,,
\end{equation*}
with $\xi(s) := \left[\bm{a}(s),\bm{b}(s),\bm{c}(s)\right] \in \mathbb{R}^{n_p\times3}$, i.e., each row in the inequality refers to a plane of the polyhedron. 
\end{subequations}
In a similar manner to \cite{kloeser2020nmpc}, we propose to embed the tunnel's bounds into $\xi$ by $n_p$ independent cubic polynomials, each with $k_{\tau}N-1$ sections:
\begin{equation}\label{eq:tunnel_constraint_2}
{}^{i}\xi(s) = \sum_{n=0}^{k_{\tau}N-1}\sum_{p=0}^{3}{}_{n}^{i}\alpha_{\,p}\, s^p,\qquad i=1,2,3
\end{equation}
where  $k_{\tau}$ is the \emph{tunnel fidelity factor} and  $i$ refers to the column of $\xi$. Coefficients ${}^{i}_{n}{\alpha}_{0,1,2,3}$ are computed by imposing continuity on first derivatives between two subsequent polynomials, $n$ and $n+1$, and using the start and final radius of the tunnel's section $n$. This yields a $C^2$ continuous constraint. To account for time-variance, all $12n_p\,(k_{\tau}N-1)$  coefficients are updated at every NMPC iteration, as denoted in \eqref{eq:tunnel_constraint}. 
\begin{comment}
Notice that the path-parametric reformulation has allowed to reduce three dimensional dynamic spatial boundaries into $n_p$ singularity-free and convex constraints on three states. 
\end{comment}
\begin{comment}
\subsection{Differential Flatness of Optimal Thrusts} 
\end{comment}

\section{EXPERIMENTS}\label{sec:experiments}
\noindent To evaluate our approach, we split the experimental analysis into three parts. First, we compare it to offline computed time-optimal trajectories in three different benchmarking tracks. Second, we test the method's ability to deal with tunnel changes outside the prediction horizon, and third, we increase the challenge with fully dynamic tunnels by also allowing modifications within the horizon.

\begin{figure*}[!t]
	\includegraphics[width=\linewidth]{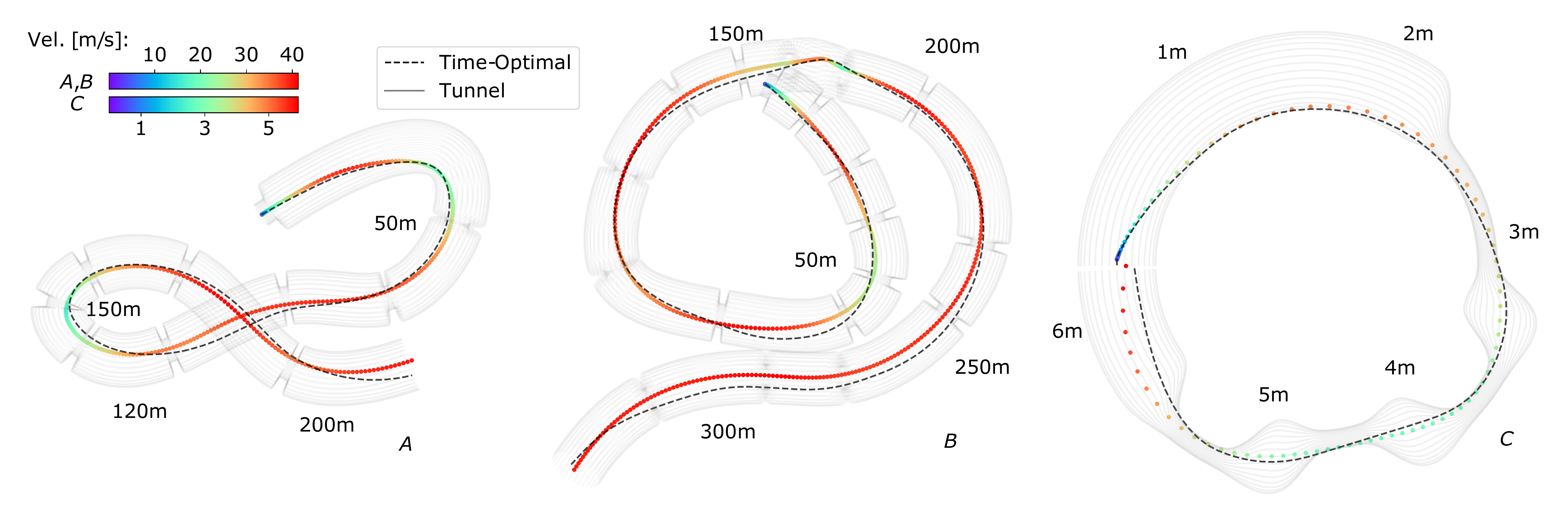}
	\vspace{-7mm}
	\caption{Top view comparison of trajectories obtained from our path-parametric NMPC-based real-time control against the time-optimal trajectory in three different tracks: A) AlphaPilot, B) AirSim, C) TubeTwist. The progress stamps might help to visualize the scale of each track. For a full three dimensional perspective see Fig. \ref{fig:trajectory_fancy}.} \label{fig:2d_trajectories}
	\vspace{-8mm}
\end{figure*}

\subsection{Experimental Setup}
\noindent We solve the NLP \eqref{eq:NLP} using the optimal control framework ACADOS \cite{verschueren2018towards} by a sequential quadratic programming (SQP) method. To account for real-time applicability, we employ the real-time iteration variant (SQP-RTI) \cite{diehl2005real,gros2020linear}, where solving for suboptimal controls enables fast enough computations. The underlying quadratic programs are solved by exploiting their multi-stage structure with HPIPM \cite{frison2020hpipm}. To integrate the dynamics, an explicit 4th-order Runge-Kutta method is used.

We choose a prediction horizon of \SI{1}{\second} with $50$ shooting nodes, resulting in a sampling time of \SI{20}{\milli\second} on a Intel Core i7-10850H notebook running Ubuntu 18.04. For all evaluations the weighting matrices are kept constant:
\begin{align*}
 &Q=\text{blkdiag}(10, 10^{-10}I_2,I_3,10^{-10}I_7, 10^{-3}I_4),\quad R= I_4\\
 &Q_N=\text{blkdiag}(10^{-10}I_3,1,10I_2,10^{-10}I_7, 10^{-3}I_4)
\end{align*}
where $I_n$ refers to an $n\times n$ identity matrix. States respective to the path coordinates and their derivatives along with rotor-thrusts are highly penalized, while the remaining diagonal values are kept small for numerical stability. 

The B-Splines corresponding to the curvature, torsion, progress-function and Frenet-Serret frame components are computed by splitting the path into \SI{30}{\centi\meter} intervals. We parametrize the tunnel as a four-sided polyhedron $n_p=4$ with a fidelity factor of $k_\tau=10$. Notice that the tunnels depicted in Figs.~\ref{fig:trajectory_fancy} and ~\ref{fig:2d_trajectories} have been computed with $n_p=32$ and are intended for visualization purposes.

Regarding the quadrotor model, we use the same racing configuration as in \cite{foehn2021time}, with thrust-to-weight ratio $6.4$, mass \SI{0.76}{\kilo\gram}, diagonal matrix inertia $[3,3,5]$ \SI{}{\gram\meter\squared}, arm-length \SI{0.17}{\meter} and rotor constant $c_\tau = 0.01$.
 
\subsection{Baseline Invariant Tunnels}
\noindent We first compare the performance of our solution to the time-optimal trajectory when navigating along three different tracks: two racetracks, from the \emph{AlphaPilot} \cite{foehn2020alphapilot} and \emph{AirSim} \cite{madaan2020airsim} challenges, and a custom-made track named \emph{TubeTwist}. Their differences on shape and scale allow for testing the versatility of the presented method. AlphaPilot and AirSim require navigating at very high speeds within wide-open spaces, while TubeTwist features flights within highly constrained, short and curled tunnels. In other words, the NMPC needs to trade-off between a local perspective required for dealing with short-term changes in narrow tunnels, such as TubeTwist, and a global perspective for approximating time-optimal behavior in long and wide tunnels, such as AlphaPilot or AirSim.

For TubeTwist we choose a reference progress velocity $\dot{s}_\text{ref}$ of \SI{10}{\meter/\second}. Its tunnel has a nominal radius of \SI{20}{\centi\meter} and contains two chicanes where it narrows down to \SI{5}{cm} and deviates up to \SI{10}{cm} from the center line. Regarding AlphaPilot and AirSim, we assign $\dot{s}_\text{ref}$ to be \SI{40}{\meter/\second} and a \SI{10}{\meter} wide tunnel has been fitted to the vertices of their respective \SI{3}{m} side-length gates. The solutions of our method when navigating along the three tunnels are presented in Figs.~\ref{fig:trajectory_fancy} and ~\ref{fig:2d_trajectories}, while the respective lap times are summarized in Table \ref{tab:deterministic}. 

\begin{table}[t!]
	\centering
	\caption{Lap times for baseline invariant tunnels shown in Figs.~\ref{fig:trajectory_fancy} and \ref{fig:2d_trajectories}.} \label{tab:deterministic}
	\begin{tabular}{|c||c|c|c|}
		\hline
		Method & AlphaPilot [s] & Airsim [s] & TubeTwist [s] \\
		\hline
		Time-Optimal & 6.38 & 9.06 & 1.23\\
		\hline
		SQP & 6.48 & 9.44 & 1.6 \\
		\hline
		SQP-RTI & 6.58 & 9.58 & 1.86 \\
		\hline
	\end{tabular}
\end{table}
\begin{comment}
The baseline lap times denoted as \emph{time-optimal} have been computed offline by solving OCP \ref{eq:OCP} in a multiple-shooting fashion. The obtained solutions represent an ideal scenario in which the update rate of thrust commands is so high that they almost become continuous, allowing for a full exploitation of the actuation, either having them at maximum or minimum values in a bang-bang manner, and thus operating at the quadrotor's theoretical physical limit. Reductions of the obtained lap times with respect to the ones in \cite{foehn2021time,song2021autonomous} are attributed to   absence of model-plant mismatch, such as drag, and assuming the quadrotor to be a point mass capable of passing through the gates by tangentially touching their borders.
\end{comment}
The \emph{time-optimal} values have been computed offline by solving OCP \ref{eq:OCP} in a multiple-shooting fashion with an interior-point-based solver \cite{wachter2006implementation}. The corresponding solutions set a theoretical lower bound that can only be matched by a full exploitation of the actuation in a bang-bang manner. Reductions with respect to the times presented in \cite{song2021autonomous} are attributed to absence of model-plant mismatch, such as drag, and the assumption that the quadrotor is a point mass capable of passing through the gates by tangentially touching their borders. 

Lap times under SQP feature the converged optimal solution of NLP \eqref{eq:NLP}. Since the penalization of thrust commands is embedded into the NLP's formulation, bang-bang controls will by definition be suboptimal, and thus its lap times will always exceed the time-optimal ones. Seeking real-time applicability, SQP-RTI relaxes the convergence condition, leading to an average  increase of \SI{0.45}{s} across all tracks with respect to the time-optimal solution. However, in contrast to the optimal approaches, an averaged computation time of \SI{21}{\milli\second} allows for its real-time implementation.

\begin{comment}
\begin{figure}[!b]
	\centering
	\includegraphics[width=\linewidth]{figs/time_vs_horizon.pdf}
	\caption{SQP-RTI lap times and average computation times depending on the quantity of shooting nodes in the prediction horizon. The dashed line depicts the time-optimal solution.}\label{fig:computation_times}
\end{figure}

Comparing SQP-RTI computation and lap times against the number of nodes in the prediction horizon, Fig.~\ref{fig:computation_times} shows that for more than $50$ nodes, lap times remain close to constant, while computation times increase linearly.
\end{comment}

\subsection{Navigating Dynamic Tunnels}
\begin{comment}
\begin{figure}
	\centering
	\includegraphics[width=\linewidth]{figs/tunnel_fidelity_small.pdf}
	\caption{SQP-RTI (N=50) lap time percentage increases and average computation times depending on the number of tunnel discretization points when accounting for dynamic spatial constraints.}\label{fig:tunnel_fidelity}
\end{figure}
\end{comment}

\begin{figure}
\begin{minipage}{.14\linewidth}
\centering
\subfloat[]{\label{fig:tunnel_randomization}\includegraphics[width=\linewidth]{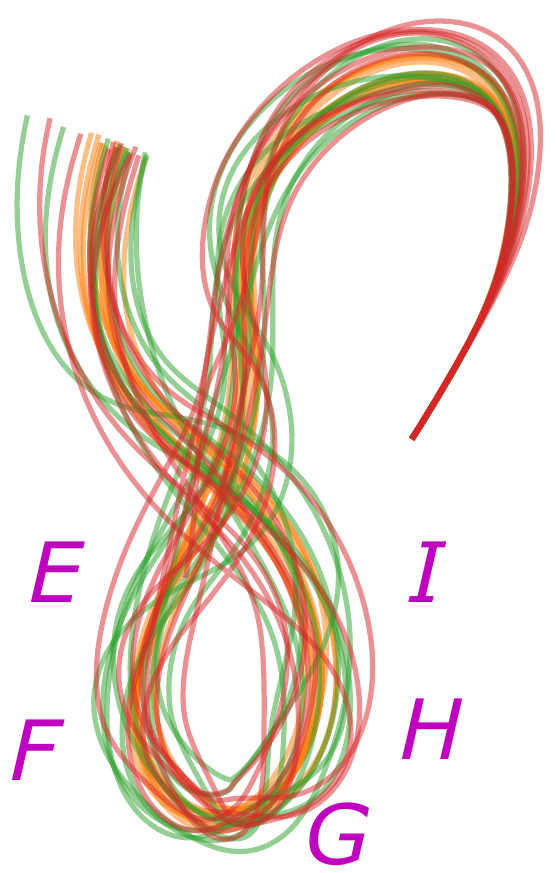}}
\end{minipage}%
\begin{minipage}{.43\linewidth}
\centering
\subfloat[]{\label{fig:hairpin}\includegraphics[width=\linewidth]{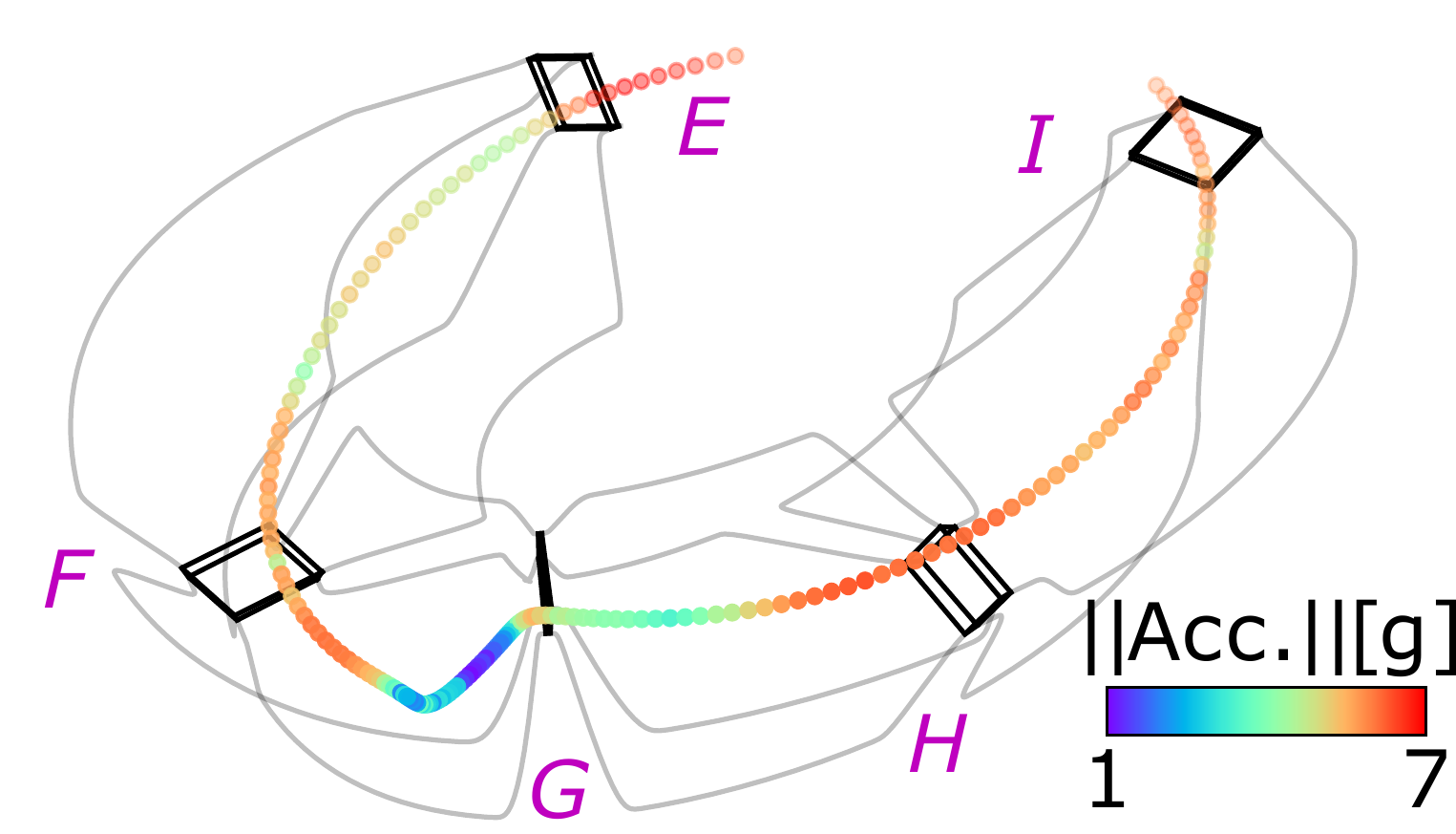}}
\end{minipage}%
\begin{minipage}{.43\linewidth}
\centering
\subfloat[]{\label{fig:hairpin_acc}\includegraphics[width=\linewidth]{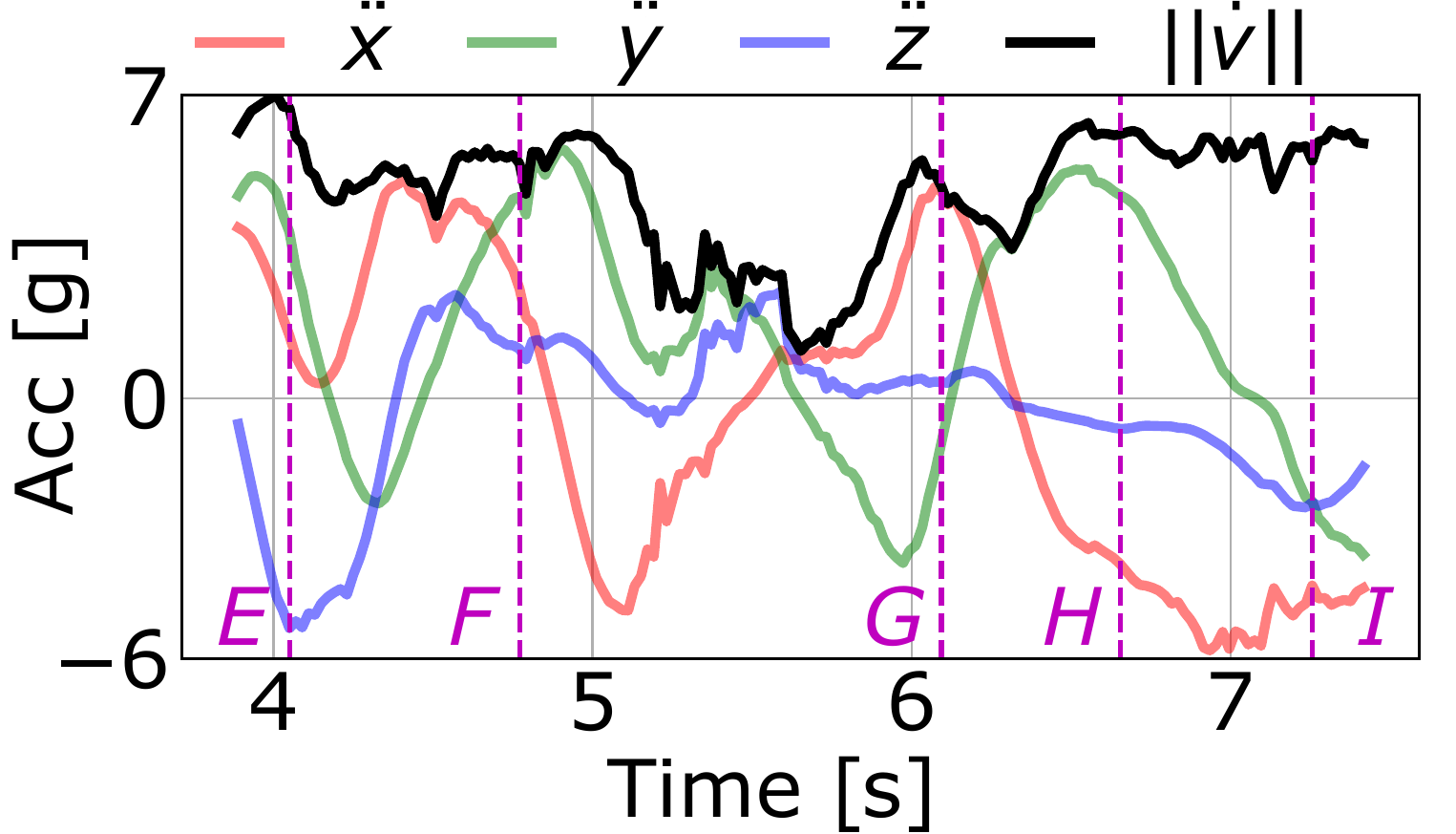}}
\end{minipage}
\caption{\textit{(a)} Top view of trajectories on different difficulty levels of dynamic AlphaPilot track, each evaluated for ten samples. From easy to difficult: orange, green, red. \textit{(b)} 3D perspective of exemplary aggressive maneuver in hairpin of maximum difficulty track. \textit{(c)} Single components and norm of acceleration while performing the maneuver.} 
\label{fig:main}%with 25\%, 75\% and 100\% of the gates in an extreme configuration
\end{figure}
\noindent To account for more realistic scenarios with changing environments and conditions, we extend the preceding subsection by navigating along tunnels that are variant outside the prediction horizon. To this end, we introduce uncertainty in the position and size of the nominal AlphaPilot gates. The resultant tunnels have been grouped into three difficulty levels, based on the percentage of gates in the extreme configuration, i.e. minimum size and maximum displacement from the center line. 

We set the minimum and maximum gate side-lengths to \SI{1.5}{\meter} and \SI{3}{\meter}, while allowing for up to \SI{4}{\meter} displacements from the center line. At an increasing order of difficulty, the levels are defined as 25\%, 75\% and 100\% and each of them is evaluated for ten different samples. As a performance metric, the corresponding time-optimal solutions have been computed. The trajectories of our method, when navigating through all randomized tracks, are depicted in Fig.~\ref{fig:tunnel_randomization}. 

The averaged lap times listed in Table \ref{tab:random} show that the gap between SQP-RTI and time-optimal solution remains consistent regardless of the difficulty level. This proves that our method is agnostic to tunnel changes beyond the prediction horizon, making it suitable for navigation in time-varying tunnels that adapt to dynamic environments.

The nature of the maximum difficulty tracks, where all gates are of the smallest size and some of them are misaligned by \SI{8}{m}, combined with the fact that the reference progress velocity has been maintained constant with respect to the original experiment at $\dot{s}_\text{ref}=40$\SI{}{\meter/\second}, results in highly challenging -- close to infeasible -- tracks. Figs.~\ref{fig:hairpin} and \ref{fig:hairpin_acc} show an exemplary maneuver that depicts this phenomenon, in which the longitudinal and lateral accelerations vary by \SI{13}{g} within \SI{3}{\second}. Given that SQP-RTI finds an approximate solution of NLP \eqref{eq:NLP}, navigating so near to the limit raises the risk of entering an infeasible state, where the only way out is to exit the tunnel by overusing the slack variables of constraint \eqref{eq:tunnel_constraint}. As summarized in the third column of Table \ref{tab:random}, this only happens for a single track at the maximum difficulty level and can be overcome by slightly lowering the reference progress velocity $\dot{s}_\text{ref}$.
\vspace{-2mm}
\begin{table}[h]
	\centering
	\caption{Averaged lap times and number of excursions for ten samples at three different difficulty levels with $\dot{s}_\text{ref}=40$\SI{}{\meter/\second}. Values inside the parenthesis refer to the difference with respect to the time-optimal solution.} \label{tab:random}
	\begin{tabular}{|M{22mm}|M{22mm}|M{22mm}|}
	    \hline
	    Diff. Level [\%] & Avg. lap time [s] &  \# Excursions [/10]\\
		\hline
		25 & 6.78 (+0.21) & 0  \\
		\hline
		75 & 7.39 (+0.38) &  0 \\
		\hline
		100  & 7.5\phantom{0} (+0.38) & 1 \\
		\hline
	\end{tabular}
\end{table}
\vspace{-2mm}
\subsection{Handling Abrupt Tunnel Changes}
\begin{figure}[!t]
	\centering
	\includegraphics[width=\linewidth]{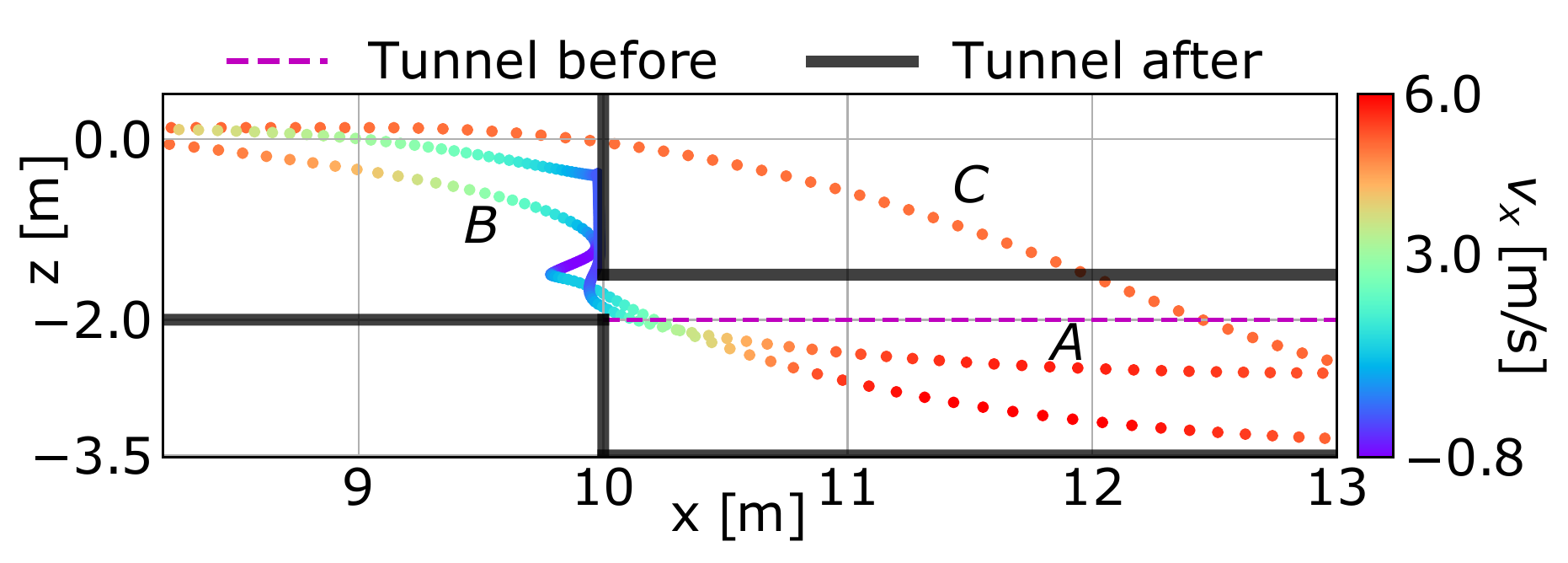}
	\caption{Trajectories when navigating at \SI{5}{\meter/\second} along a \SI{2}{\meter} wide and \SI{20}{\meter} long straight tunnel that is abruptly modified at three different locations of the prediction horizon: A,B,C for \SI{5}{\meter}, \SI{3}{\meter} and \SI{1}{\meter}. The upper tunnel bound of the first section ($x<$\SI{10}{\meter}) is not visible. The dashed line represents the original second section, while the final tunnel shows in black.} \label{fig:abrupt_changes}
\end{figure}
\noindent Agile navigation within highly dynamic environments benefits from the ability to modify the spatial bounds within the prediction horizon. Therefore, in this subsection we study the robustness of our approach when abruptly altering the tunnel at different locations along the horizon. 

To this end, we design an experiment in which the quadrotor navigates along a \SI{2}{\meter} wide and \SI{20}{\meter} long straight tunnel that suddenly shrinks to \SI{1}{\meter} in width and vertically displaces by \SI{2.5}{\meter}, leaving a \SI{50}{\centi\meter} transition gap. The abrupt change is conducted in three different locations evenly distributed along the prediction horizon while navigating at \SI{5}{\meter/\second}, at a distance of \SI{1}{\meter}, \SI{3}{\meter} and \SI{5}{\meter} from the quadrotor. Resultant trajectories are depicted in Fig.~\ref{fig:abrupt_changes} and the corresponding navigation times are summarized in Table \ref{tab:abrupt_changes}.

For case \emph{A}, where the change occurs at the last node of the horizon at a distance of \SI{5}{\meter}, the quadrotor has enough time to plan a conservative maneuver -- stop and descent vertically -- resulting in a safe passage through the narrow gap. In scenario \emph{B}, the tunnel modification is moved to the center of the horizon, decreasing the reaction distance to \SI{3}{m} and forcing the quadrotor to come to a complete stop and conduct a reverse motion. Finally, in \emph{C} the tunnel is altered just \SI{1}{m} in front of the quadrotor, making it impossible to anticipate for the abrupt change and obliging it to skip the transition.

These results suggest that our method can account for aggressive tunnel modifications within the prediction horizon. When doing so, it has demonstrated the ability to perform not only stop-and-go maneuvers, but also backward motions. Moreover, when confronted with modification excesses, it has shown itself capable of returning to the tunnel.

\vspace{-2mm}
\begin{table}[h]
	\centering
	\caption{Lap times and excursions when abruptly modifying the tunnel within the horizon. Second column values refer to the prediction node and distance from the quadrotor at which the tunnel change occurs.} \label{tab:abrupt_changes} 
	\begin{tabular}{|M{7mm}|M{35mm}|M{12mm}|M{12mm}|}
	    \hline
	    Label & Abrupt change location \newline Distance [m] -- Node [/50] & Lap time [s] & No excursion\\
		\hline
		A & 5\hspace{4mm}--\hspace{4mm}50 & 7.66 & \cmark  \\
		\hline
		B & 3\hspace{4mm}--\hspace{4mm}30 & 5.3\phantom{0} &  \cmark \\
		\hline
		C & 1\hspace{4mm}--\hspace{4mm}10  & 4.38 & \xmark \\
		\hline
	\end{tabular}
\end{table}
%\vspace{-3mm}
\section{CONCLUSION}\label{sec:conclusion}
\noindent In this work, we proposed a real-time control for quadrotors, capable of approximating time-optimality within variant spatial constraints. For this purpose, a path-parametric reformulation of the dynamics has allowed for efficiently embedding the spatial properties into an NMPC-based control scheme. The presented solution has been validated in three different tracks of varying shapes and sizes, each with its own set of challenges. Experimental results demonstrate that our solution not only converges to time-optimality while navigating at high speeds -- over \SI{40}{\meter/\second} -- and performing aggressive maneuvers -- up to \SI{7}{g} --, but it also is eligible to stop or reverse if spatial bounds are abruptly modified. Lastly, the fact that the controller's tuning was kept constant throughout all evaluations, emphasizes its applicability and versatility to a wide range of scenarios and dynamic regimes.

% REFERENCES
%\addtolength{\textheight}{-12cm}   % Balance the column lengths on the last page of the document manually.
\clearpage
\newcommand{\BIBdecl}{\setlength{\itemsep}{0.45 em}} %spacing between references
\bibliographystyle{IEEEtran}
\bibliography{time_optimal_tunnel_following_for_quadrotors}%IEEEabrv

\begin{comment}
% APPENDIX
\clearpage
\appendix
\subsection{Random tracks}
\begin{table}[h]
	\centering
	\caption{SQP-RTI: Lap times for ten samples at three different difficulty levels of AlphaPilot.} 
	\begin{tabular}{|c||c|c|c|c|c|c|c|c|c|c|c|}
	    \hline
	    {\multirow{2}{*}{Random. Level (RL)}} & \multicolumn{10}{c|}{Lap Time [s] - AlphaPilot}  & Mean\\
	    \cline{2-12}
		& 1 & 2 & 3 & 4 & 5 & 6 & 7 & 8 & 9 & 10 &\\
		\hline
		0.25 & 6.7& 6.9& 6.86& 6.72& 6.76& 6.74& 6.74& 6.74& 6.94& 6.72 &6.78\\
		\hline
		0.75 & 6.92&  7.34& 8.08& 7.42& 7.68& 7.2& 7.76& 7.8&6.58 &7.12 &7.39\\
		%\hline
		 %1 (slack 1e4 and 0.1) & 6.5 & 7.4 & 7.26 & DNF-S & 7.5 & DNF-S & 7.3& 8.08& 7.28&8.34 &7.45 \\
		\hline
		1 & 6.52& 7.42 & 7.22 & DNF-S & 7.46 &7.76 &7.42 &8.32 &7.24 &8.16 &7.5 \\
		\hline
	\end{tabular}
\end{table}

\begin{table}[h]
	\centering
	\caption{Offline multiple shooting: Lap times for ten samples at three different difficulty levels of AlphaPilot.} 
	\begin{tabular}{|c||c|c|c|c|c|c|c|c|c|c|c|}
	    \hline
	    {\multirow{2}{*}{Random. Level (RL)}} & \multicolumn{10}{c|}{Lap Time [s] - AlphaPilot}  & Mean\\
	    \cline{2-12}
		& 1 & 2 & 3 & 4 & 5 & 6 & 7 & 8 & 9 & 10 & \\
		\hline
		0.25 & 6.67 & 6.58 & 6.59 & 6.55 & 6.48 & 6.51 & 6.51 & 6.57 & 6.57 & 6.56 & 6.57\\
		\hline
		0.75 & 6.69& 7.04& 7.55& 7.17&7.16 &7.02 &7.08 &7.31 &6.3 & 6.75 &7.02   \\
		\hline
		1 & 6.4& 6.88& 6.9& 7.12 & 7.31 & 7.72 (LC) & 6.88& 7.58&6.91 &8.05& 7.12\\
		\hline
	\end{tabular}
\end{table}
\end{comment}
\end{document}